\documentclass[11pt,a4paper]{article}
\usepackage{times,latexsym}
\usepackage{url}
\usepackage[T1]{fontenc}
\usepackage[acceptedWithA]{tacl2021v1}
\usepackage{amsmath,amssymb}
\usepackage{booktabs}
\usepackage{graphicx}
\usepackage{multirow}
\usepackage{xspace}
\usepackage{microtype}
\usepackage{array}
\usepackage{makecell}

\newcommand{\full}{\textsc{FullKV}\xspace}

\newcommand{\risk}{\mathcal{R}}
\newcommand{\E}{\mathbb{E}}
\newcommand{\Prob}{\mathbb{P}}

\title{Risk-Controlled KV-Cache Eviction: From Memory Budgets to Risk Targets}
\author{
Beomgu Kang, SoJin Yun, Hojoon Kim, Hyunseok Seo\textsuperscript{*} \\
Department of Artificial Intelligence, Korea University, Republic of Korea \\
}
\date{}

\begin{document}
\maketitle

\begingroup
\renewcommand{\thefootnote}{}
\footnotetext{\textsuperscript{*}Corresponding author: seoh@korea.ac.kr}
\addtocounter{footnote}{-1}
\endgroup

\begin{abstract}
KV-cache eviction is typically evaluated through average quality–memory trade-offs, yet a small average loss can hide requests whose utility degrades materially. We reformulate eviction as a deployment risk-control problem: a material degradation occurs when eviction lowers task utility by more than a deployment-specified tolerance relative to full-KV inference on the same request, and deployment risk is the population frequency of such events. Given a reliability contract specifying a target risk level and confidence requirement, we use a compressor-agnostic post-hoc certification procedure to select a retention policy from calibration data with a finite-sample guarantee, falling back to full KV when no compressed policy is certified. Across multiple eviction methods, Llama and Mistral models, and LongBench and RULER-32K, the same contract supports substantially different levels of eviction: on Llama, it certifies SnapKV at 75\% retention on LongBench but no tested compressed policy on RULER-32K, triggering full-KV fallback. Policies with empirical degradation rates below the 5\% target can still fail finite-sample certification; on Llama LongBench, empirical thresholding selects uncertified policies that retain 5--10 percentage points less cache across fixed-budget methods. The proposed framework converts a deployment-level reliability requirement into a KV-memory operating point.
\end{abstract}

\section{Introduction}
\label{sec:intro}
Large language models (LLMs) increasingly support extended context windows, enabling long-context workloads such as multi-document question answering, long-document summarization, and code completion~\citep{bai2024longbench}. However, efficiently serving these long-context LLMs remains challenging due to the growing memory cost of the key-value (KV) cache. During autoregressive decoding, the KV cache stores the key and value states of previously processed tokens to avoid redundant recomputation and accelerate token generation. As the sequence length and the number of active requests increase, the KV cache grows accordingly and can become a major memory bottleneck. This challenge has motivated a growing body of work on KV-cache eviction~\citep{zhang2023h2o,li2024snapkv,feng2025adakv,cai2025pyramidkv,zhou2025dynamickv,feng2026defensivekv,feng2026criticalkv}. These methods primarily focus on which KV states to retain under a limited memory budget while preserving model quality.

\begin{figure}[t]
    \centering
    \includegraphics[width=\columnwidth]{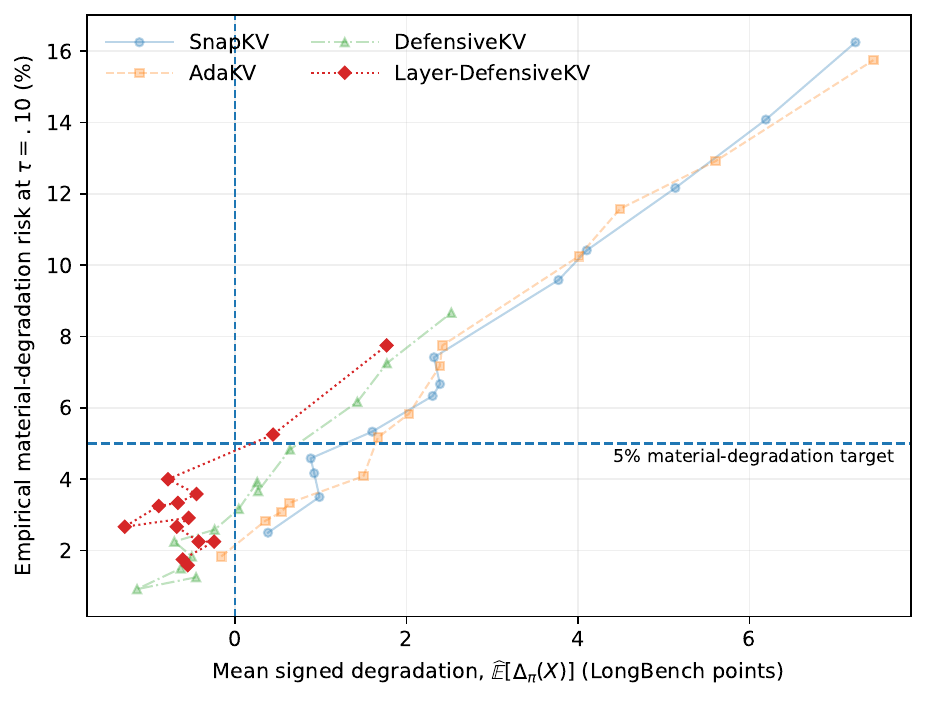}
    \caption{
    Mean degradation versus request-level risk for Llama-3.1-8B on LongBench.
    Similar average degradation can correspond to different
    material-degradation risks under $\tau=.10$.
    Lines connect candidate policies from conservative to aggressive eviction,
    and dashed lines indicate zero mean degradation and the target
    $\epsilon=.05$.
    }
    \label{fig:mean_tail_intro}
\end{figure}

Deployment introduces a complementary question: \emph{how aggressive can an eviction policy be while remaining reliable at the request level?} Good average benchmark performance does not guarantee that individual requests are reliably preserved.  Let $U_{\mathrm{full}}(X)$ denote task utility under full KV and $U_\pi(X)$ the utility under an eviction policy $\pi$. Define the signed eviction-induced degradation
\begin{equation}
    \Delta_\pi(X)=U_{\mathrm{full}}(X)-U_\pi(X).
\end{equation}
A small average degradation does not imply that large request-level
degradations are rare. In particular, a small value of
$\E[\Delta_\pi(X)]$ does not imply
\begin{equation}
    \Prob[\Delta_\pi(X)>\tau]\le\epsilon.
    \label{eq:mean_not_risk}
\end{equation}
The mean captures aggregate quality, whereas the exceedance probability
captures how often eviction causes a drop larger than a deployment
tolerance. Figure~\ref{fig:mean_tail_intro} illustrates this distinction
under our main materiality threshold $\tau=.10$. Across eviction methods
and retention ratios, configurations with similar mean degradation can
exhibit different material-degradation risks. This motivates controlling
the exceedance risk directly rather than selecting an eviction budget
from average performance alone.

We address this deployment problem through finite-sample risk control. Building on Learn-then-Test (LTT)~\citep{angelopoulos2025ltt}, we use an independent calibration set to determine how aggressively the KV cache may be evicted while controlling the frequency of request-level utility degradations that exceed a deployment-specified tolerance. We refer to such tolerance-violating cases as \emph{material degradation events}. Our procedure is a compressor-agnostic calibration wrapper: it does not modify the underlying eviction algorithm and can be applied to any pre-specified family of policies. 

Given a degradation tolerance and a reliability requirement,
fixed-sequence testing evaluates a pre-specified sequence of policies and
returns the most aggressive policy certified before the first failure,
while falling back to full KV if no eviction policy can be certified. We evaluate four fixed-budget methods with Llama and
Mistral on LongBench~\citep{bai2024longbench} and with Llama on
RULER-32K~\citep{hsieh2024ruler}: SnapKV~\citep{li2024snapkv},
AdaKV~\citep{feng2025adakv}, DefensiveKV, and Layer-DefensiveKV
~\citep{feng2026defensivekv}. We additionally evaluate the request-adaptive
ReFreeKV~\citep{ni2025refreekv} with Llama on LongBench. These settings allow
us to study risk-controlled deployment across different forms of KV-memory
allocation.

The main contributions of our work are summarized as follows:
\begin{itemize}

\item \textbf{Reliability-first formulation of KV-cache eviction:}
We formulate KV-cache eviction as a request-level risk-control problem,
where a reliability contract specifies the tolerated utility loss and its
allowed population frequency relative to paired full-KV inference.

\item \textbf{Compressor-agnostic finite-sample certification:}
Building on Learn-then-Test, we instantiate a fixed-sequence procedure that
selects the most aggressive retention policy, with full-KV fallback, without modifying the underlying eviction
algorithm. Our task-stratified calibration design provides a finite-sample
guarantee for the declared task-balanced population.

\item \textbf{Empirical consequences of certification:}
Certified operating points vary substantially across models, compressors,
and long-context benchmarks under the same reliability contract. On Llama
LongBench, finite-sample certification selects $5$--$10$ percentage points
more retention than empirical thresholding, even when the empirical calibration risk does not exceed $5\%$.
\end{itemize}

\section{Related Work}
\label{sec:related}

\subsection{KV-cache eviction}
KV-cache eviction reduces memory by retaining a subset of cached states
while discarding those deemed less important. H2O retains recent tokens
together with attention heavy hitters~\citep{zhang2023h2o}, while SnapKV
uses an observation window to identify prompt positions likely to remain
useful during generation~\citep{li2024snapkv}. AdaKV allocates a global
budget non-uniformly across attention heads~\citep{feng2025adakv},
whereas PyramidKV and DynamicKV adapt retention across layers
~\citep{cai2025pyramidkv,zhou2025dynamickv}. 
DefensiveKV further improves robustness to shifts in token importance through
defensive aggregation, while its Layer-DefensiveKV extension additionally
allocates the KV budget across layers~\citep{feng2026defensivekv}. 
\citet{chen2026pitfalls} show that KV-cache compression
can disproportionately degrade individual instructions in
multi-instruction prompts, using system-prompt leakage as a case
study of instruction-following failure. They mitigate these failures
by protecting selected tokens or balancing eviction across
instruction spans.

Beyond adapting how a fixed budget is distributed within a request, recent work has explored adapting the total KV budget across requests.
ReFreeKV uses an input-dependent signal to determine request-specific
retention~\citep{ni2025refreekv}, while CompilerKV uses prompt-level
signals such as attention entropy and local perplexity to adapt compression
aggressiveness through offline-compiled thresholds
~\citep{yang2026compilerkv}. These methods adapt retention to the input but
do not report finite-sample control of the population frequency of
downstream material degradation.

Recent work has introduced formal guarantees for KV compression.
\citet{xie2026errorcert} derives per-step attention-output error
certificates by replacing deterministic eviction with a randomized sampling
design, while WitCert develops runtime
attention-fidelity bounds and fallback gating for KV-cache quantization
~\citep{wei2026witcert}. Both certify how faithfully attention is preserved, while downstream task quality is evaluated empirically. We instead certify a bound on the population frequency of request-level task degradations relative to full-KV inference, without modifying the underlying eviction mechanism.

\subsection{Distribution-free and quantile risk control for inference}

Distribution-free risk-controlling prediction sets provide
finite-sample guarantees on population risk using calibration
data~\citep{bates2021distribution}.
Learn-then-Test (LTT) casts candidate configurations as a family
of statistical hypotheses, constructs valid p-values from
calibration outcomes, and applies multiple-testing procedures
to obtain risk guarantees for selected configurations
~\citep{angelopoulos2025ltt}.
Quantile Risk Control highlights that expected loss may obscure the
high-loss tail of a distribution~\citep{snell2023quantile}. Our material-degradation event asks a related tail question through a fixed-threshold
exceedance probability, $\Prob[\Delta_\pi(X)>\tau]\le\epsilon$;
finite-sample certification is performed using LTT.

A closely related line of work is risk-controlled adaptive inference. CALM dynamically allocates decoder computation through \emph{early exiting} and calibrates stopping thresholds to preserve generation quality~\citep{schuster2022calm}. Fast yet Safe explicitly combines early exiting with post-hoc distribution-free risk control across vision and language tasks~\citep{jazbec2024fast}. \citet{wynn2026controlling} use related machinery to control degradation from corrupted context relative to a zero-shot baseline. Conformal Thinking reframes adaptive reasoning as a compute-allocation problem, calibrating stopping rules to limit risk while reducing reasoning tokens~\citep{wang2026conformal}. Collectively, these works illustrate how risk-control frameworks can calibrate resource-saving inference decisions from a reliability requirement rather than solely from a preset resource budget.

These methods adapt how much computation to allocate to an input, through choices such as decoder depth or reasoning length. KV-cache eviction instead requires calibrating how aggressively cached context can be discarded: candidate policies may vary a scalar retention ratio, redistribute a global budget across heads or layers, or adapt retention across requests. We define risk as the population frequency of material degradation relative to paired full-KV inference and instantiate a task-stratified finite-sample certification procedure over these pre-specified policy families, without modifying the underlying compressor.

\section{Risk-Controlled KV Deployment}
\label{sec:method}

\subsection{Risk-Controlled KV Eviction Formulation}

Let $X$ denote a request drawn from a target population
$\mathcal{P}$, defined below as the uniform mixture over the
evaluated tasks.
An eviction policy $\pi$ determines the KV representation retained
for generation.
It may use a fixed retention ratio, allocate a global KV budget
nonuniformly across heads or layers, or adapt retention to each
request.

Let $U_\pi(X)\in[0,1]$ and $U_{\mathrm{full}}(X)$ denote the
normalized downstream utilities under $\pi$ and full KV,
respectively, evaluated using the same model, prompt, generation
settings, and evaluator.
Define the signed degradation ($\Delta_\pi$) and material-degradation indicator ($V_\tau$) as
\begin{equation}
    \Delta_\pi(X)
    =
    U_{\mathrm{full}}(X)-U_\pi(X),
    \label{eq:degradation}
\end{equation}
\begin{equation}
    V_\tau(X;\pi)
    =
    \mathbf{1}\!\left\{\Delta_\pi(X)>\tau\right\},
    \label{eq:indicator}
\end{equation}
where $\tau\in(0,1)$ is the material-degradation tolerance.
The deployment risk is
\begin{equation}
    \risk_\tau(\pi)
    =
    \E_{X\sim\mathcal{P}}
    \left[V_\tau(X;\pi)\right].
    \label{eq:risk}
\end{equation}
This risk measures degradation relative to full KV rather than
absolute task failure.
Utility improvements on some requests do not offset material
degradations on others.

The reliability contract requires
\begin{equation}
    \risk_\tau(\pi)\le\epsilon,
    \label{eq:main_contract}
\end{equation}
where $\epsilon$ is the maximum allowable frequency of material
degradation.
Our main setting, $(\tau,\epsilon)=(0.10,0.05)$, permits utility
drops greater than $0.10$ on at most $5\%$ of requests under
$\mathcal{P}$.
For the policy $\widehat{\pi}$ returned by calibration, including
the full-KV fallback, we seek the finite-sample guarantee
\begin{equation}
    \Pr_{\mathrm{CAL}}
    \left\{
        \risk_\tau(\widehat{\pi})>\epsilon
    \right\}
    \le\delta,
    \label{eq:selected_policy_guarantee}
\end{equation}
where probability is over the calibration sample (CAL).
We use $\delta=0.05$ in the main setting.
Here, $\epsilon$ bounds the population frequency of a request-level event, whereas $\delta$ bounds the probability, over calibration sampling, of returning a policy that violates that population bound. Neither is a guarantee for each individual request. The finite-sample statement is conditional on the sampling and pre-specification assumptions below.

For a fixed-budget policy, let $b(\pi)\in(0,1]$ denote the
retained-cache fraction.
The ideal deployment objective is
\begin{equation}
    \min_{\pi\in\Pi} b(\pi)
    \quad\text{s.t.}\quad
    \risk_\tau(\pi)\le\epsilon.
    \label{eq:optimization}
\end{equation}
For request-adaptive policies, the objective can instead use
$\E_{X\sim\mathcal{P}}[b_\pi(X)]$.
Rather than directly solving \eqref{eq:optimization}, we select
from a finite, pre-specified policy sequence using statistical
certification, with full KV as the fallback.
This procedure guarantees risk control, not global resource
optimality.

\subsection{Finite-Sample Certification with Fixed-Sequence Testing}
\label{sec:certification}

We instantiate Learn-then-Test (LTT)
~\citep{angelopoulos2025ltt} by treating each candidate eviction
policy as a risk hypothesis and applying fixed-sequence testing
to valid calibration p-values.
Our task-stratified calibration design uses the binomial
comparison below to establish p-value validity while allowing
violation probabilities to differ across task strata.

\paragraph{Candidate sequence.}
For fixed-budget compressors, we pre-specify the retention sequence
\[
    (b_1,\ldots,b_m)
    =
    (0.80,\,0.75,\,\ldots,\,0.20),
\]
where $\pi_j:=\pi(b_j)$ and $b_1>\cdots>b_m$ orders candidates
from conservative to aggressive compression.
The shared grid enables comparisons across compressors at matched
retention budgets without compressor-specific grid tuning on CAL.

For every confirmatory model--compressor--benchmark sequence, the
candidate policies, their order, and $(\tau,\epsilon,\delta)$ are fixed
before its CAL outcomes are inspected; any policy fitting or
compressor-specific tuning uses separate data. Candidates may be evaluated
sequentially, stopping at the first non-certification. When post-stop
candidates are additionally evaluated for descriptive curves, their
outcomes do not enter confirmatory selection or alter the stopping rule.
TEST is not used to choose the certified operating point.

\paragraph{Calibration design and target population.}
Let $\mathcal{P}_h$ denote the request distribution for task
$h\in\{1,\ldots,H\}$.
Our target population is the task-uniform mixture
\begin{equation}
    \mathcal{P}
    =
    \frac{1}{H}\sum_{h=1}^{H}\mathcal{P}_h.
    \label{eq:target_mixture}
\end{equation}
CAL contains $n/H$ observations from each task.
We assume that calibration requests are independently sampled from
their corresponding task distributions $\mathcal{P}_h$.

For a fixed candidate $\pi_j$, let
$q_{ij}=\Pr\{V_\tau(X_i;\pi_j)=1\}$.
Equal task allocation implies
\begin{equation}
    \bar q_j
    :=
    \frac{1}{n}\sum_{i=1}^{n}q_{ij}
    =
    \frac{1}{H}\sum_{h=1}^{H}
    \E_{X\sim\mathcal{P}_h}[V_\tau(X;\pi_j)]
    =
    \risk_\tau(\pi_j).
    \label{eq:calibration_risk_alignment}
\end{equation}
The guarantee therefore applies to task-uniform mixture risk,
not per-task risk or risk under different deployment task
proportions or within-task distributions.

\paragraph{Finite-sample risk test.}
For candidate $\pi_j$, let
\begin{equation}
    K_j
    =
    \sum_{i=1}^{n}V_\tau(X_i;\pi_j)
    \label{eq:violation_count}
\end{equation}
be the CAL violation count, with empirical risk
$\widehat{\risk}_{\tau,j}=K_j/n$.
We test
\begin{equation}
    H_j:\risk_\tau(\pi_j)\ge\epsilon
    \quad\text{vs.}\quad
    H_j^{A}:\risk_\tau(\pi_j)<\epsilon.
    \label{eq:hypothesis}
\end{equation}
We conservatively include equality in the null, so certification
requires evidence that risk is strictly below the target.

The violation indicators ($V_\tau$) are independent Bernoulli variables
whose probabilities may differ across tasks.
Under $H_j$, \eqref{eq:calibration_risk_alignment} gives
$\bar q_j\ge\epsilon$.
For integer $k\le\lfloor n\epsilon\rfloor-1$, Hoeffding's
comparison theorem~\citep{hoeffding1956distribution} and
monotonicity of the binomial lower tail imply
\begin{equation}
    \begin{aligned}
        \Pr(K_j\le k)
        &\le
        \Pr\{\mathrm{Bin}(n,\bar q_j)\le k\}\\
        &\le
        \Pr\{\mathrm{Bin}(n,\epsilon)\le k\}.
    \end{aligned}
    \label{eq:binomial_envelope}
\end{equation}
Thus,
\begin{equation}
    p_j^{\mathrm{env}}
    =
    \begin{cases}
        \Pr\{\mathrm{Bin}(n,\epsilon)\le K_j\},
        & K_j\le\lfloor n\epsilon\rfloor-1,\\
        1, & \text{otherwise}
    \end{cases}
    \label{eq:penv}
\end{equation}
is a valid, possibly conservative p-value:
$\Pr\{p_j^{\mathrm{env}}\le\alpha\}\le\alpha$ under $H_j$
for every $\alpha\in[0,1]$.
A candidate reached by fixed-sequence testing is certified
when $p_j^{\mathrm{env}}\le\delta$.

For $n=1200$, $\epsilon=0.05$, and $\delta=0.05$, the largest
certifiable violation count is $K_{\max}=47$, corresponding to
an empirical risk of $3.92\%$.
This is stricter than the nominal $5\%$ target because of
finite-sample uncertainty.

\paragraph{Fixed-sequence selection and fallback.}
Candidates are tested in the pre-specified order, from conservative
to increasingly aggressive compression.
Testing continues while $p_j^{\mathrm{env}}\le\delta$ and stops at
the first non-rejection, returning the last certified candidate.
If all candidates are certified, the procedure returns $\pi_m$;
if the first candidate is not certified, it returns \full.
The fallback is identified with the paired full-KV reference, so that
$\Delta_{\mathrm{full}}(X)=0$ and
$\risk_\tau(\mathrm{full})=0$ by definition.

Within a pre-specified sequence, fixed-sequence testing controls the
probability of any false certification at level $\delta$
~\citep{angelopoulos2025ltt}: rejecting any true null requires rejecting the
first true null in that sequence, whose rejection probability is at most
$\delta$. Together with the full-KV fallback, this establishes
\eqref{eq:selected_policy_guarantee}. The result permits arbitrary dependence
among candidate p-values and does not require risk to be monotone in
retention.
The returned policy is the last policy certified before the procedure stops, not necessarily the lowest-retention candidate that could pass under another testing procedure. When risk is non-monotone, stopping at the first non-rejection can exclude later candidates that would pass their individual tests.

The guarantee is sequence-specific. We apply it separately to each
model--compressor--benchmark sequence and do not claim a simultaneous
$1-\delta$ guarantee across all reported sequences. Selecting among multiple
sequence outputs using their CAL outcomes requires additional multiplicity
control or fresh calibration data.

\subsection{Request-Adaptive Policies}
\label{sec:adaptive_method}

The same framework applies to request-adaptive compressors.
For example, ReFreeKV uses a threshold $T$ to determine a
request-specific retained fraction $b_T(X)$
~\citep{ni2025refreekv}.
Each threshold therefore defines a policy $\pi_T$, despite
variable memory usage across requests.

We fix the ordered threshold sequence
\[
\begin{aligned}
(T_1,\ldots,T_{10})={}&(.0005,.001,.002,.003,.005,\\
                       &\ .0075,.01,.015,.02,.03)
\end{aligned}
\]
independently of CAL outcomes, following the compressor's intended progression
from conservative to aggressive compression.
For each $\pi_{T_j}$, we compute the same degradation and violation
indicators and apply the finite-sample test and fixed-sequence
selection rule, including the full-KV fallback.

For the selected policy, we estimate its mean retained fraction
$\E_{X\sim\mathcal{P}}[b_T(X)]$ overall and within each task on TEST using
equal task weights.

\section{Experimental Setup}
\label{sec:experiments}

\subsection{Models and eviction methods}

We evaluate two open-weight LLMs:
Llama-3.1-8B-Instruct~\citep{grattafiori2024llama3} and
Mistral-7B-Instruct-v0.3~\citep{jiang2023mistral}.
Our primary LongBench evaluation considers four representative fixed-budget
KV-cache eviction methods: SnapKV~\citep{li2024snapkv},
AdaKV~\citep{feng2025adakv}, DefensiveKV, and Layer-DefensiveKV
~\citep{feng2026defensivekv}. Layer-DefensiveKV extends DefensiveKV with
layer-wise budget allocation, allowing different layers to retain different
amounts of KV cache. We additionally evaluate ReFreeKV~\citep{ni2025refreekv}
as a request-adaptive method whose realized retention varies across requests.
The reported experiments were run on NVIDIA RTX A6000 (48 GB) GPUs.
Fixed-budget runs use FlashAttention-2~\citep{dao2024flashattention}, whereas the ReFreeKV implementation
uses eager attention.

\subsection{Benchmarks and evaluation populations}

\paragraph{LongBench Benchmark.}
LongBench~\citep{bai2024longbench} contains heterogeneous long-context
tasks spanning question answering, summarization, retrieval,
classification, and code, with task-specific evaluation metrics.
We define the evaluation population as a task-balanced mixture so that
tasks with more available examples do not dominate the aggregate
deployment risk. We include the 12 tasks with at least 200 distinct contexts
in the preprocessed corpus: 2WikiMQA, GovReport, HotpotQA, LCC, MultiNews,
MuSiQue, PassageCount, PassageRetrieval-en, RepoBench-P, SAMSum, TREC, and
TriviaQA. This inclusion rule and task set were fixed before CAL outcomes
were inspected.

For each task, we uniformly sample 200 eligible requests without replacement using the fixed random seed 20260904. We then randomly partition them into disjoint 100-request calibration (CAL) and held-out test (TEST) splits, yielding \(n=1200\) requests per split. For the finite-sample analysis, requests within each task stratum are modeled as independent draws from the corresponding task distribution \(\mathcal{P}_h\). The resulting certificate therefore applies to the aggregate task-balanced population, not to each task individually; per-task TEST risks are reported only as descriptive diagnostics.

ReFreeKV is evaluated separately under its implementation's 8K protocol. The
input budget is 8192 tokens minus the task-specific generation limit and a
20-token margin; longer prompts retain the first and last halves of the
available budget. Its prompt formatting also differs from the fixed-budget
protocol on some tasks. ReFreeKV and its full-KV reference use identical
processed inputs and generation settings. This experiment therefore assesses
adaptive-policy calibration only within that matched protocol, and its
certificate does not cover preprocessing-induced degradation. We do not make
direct cross-protocol comparisons of absolute utility or retention efficiency.

\paragraph{RULER Benchmark.}
RULER~\citep{hsieh2024ruler} provides 13 controlled long-context tasks
covering needle-in-a-haystack retrieval, variable tracking, word
extraction, and question answering. We evaluate all 13 tasks at a
32K context length and define the RULER population analogously as a
uniform mixture over tasks.

CAL and TEST each contain 100 requests per task, yielding $n=1300$
requests per split. For this sample size and
$(\epsilon,\delta)=(.05,.05)$, the largest certifiable violation count is
$K_{\max}=51$ (3.92\% empirical CAL risk). We independently apply the same
calibration and fixed-sequence certification procedure to this RULER population.
As with LongBench, the resulting certificate is an aggregate guarantee
over the task-balanced population rather than a simultaneous guarantee
for each individual task.

\subsection{Reliability Contract and Experimental Details}

Unless stated otherwise, each confirmatory model--compressor--benchmark
sequence uses $(\tau,\epsilon,\delta)=(0.10,0.05,0.05)$. Thus, with
probability at least $0.95$ over calibration sampling, the selected policy
has population material-degradation risk at most $0.05$ under the declared
task-balanced population. This guarantee applies separately to each
confirmatory sequence, rather than simultaneously across all methods,
models, and benchmarks, and does not require the finite-sample TEST
violation rate to be below $0.05$.

For both benchmarks, we normalize the official example-level metric to
$[0,1]$ before computing paired degradation. A material-degradation event is
then determined from the paired compressed and full-KV utilities according
to Eq.~\ref{eq:indicator}.

For fixed-budget experiments, we use the prompt, answer prefix, and
task-specific generation limit stored with each preprocessed benchmark
request. We
instantiate each eviction method with the default hyperparameters of its
evaluated implementation and vary only the retained fraction. All four
methods use an observation window of 32 and kernel size 5, and AdaKV
additionally uses $\alpha=0.2$. The retained fraction is defined relative to
the processed prefill sequence length, with the 32-token observation window
counted within the stated budget. Integer budgets and method-specific
allocation across heads and layers follow the evaluated implementations.
All generations use deterministic greedy decoding, eliminating decoding-sampling randomness from the paired full-KV and compressed evaluations.

\subsection{Calibration and Held-out Evaluation Protocol}

Policy selection is performed exclusively on CAL. Candidates are tested in
the pre-specified conservative-to-aggressive order until the first
non-certification. Post-stop candidates are evaluated only for the
descriptive Llama LongBench analyses and do not affect selection.

After the CAL-selected policy is frozen, TEST is used only for descriptive
held-out evaluation against its paired full-KV baseline. No TEST outcome is
used to modify the candidate sequence, certification decision, or selected
operating point. LongBench and RULER are calibrated independently.

\section{Results}
\label{sec:results}

\subsection{LongBench Results}
\label{sec:longbench_results}

\subsubsection{Selected Operating Points and Held-out Evaluation}
\label{sec:main_longbench}

Table~\ref{tab:main_lb} reports the retained fraction of the policy selected by fixed-sequence certification for each model--compressor sequence. Across both Llama-3.1-8B and Mistral-7B, the certified operating points vary substantially across eviction methods, with the DefensiveKV variants supporting markedly lower retained fractions than SnapKV and AdaKV. The CAL-selected retention ratios differ by model. In particular,
Layer-DefensiveKV reaches lower retention than DefensiveKV on Llama
($0.35$ versus $0.40$), whereas the order reverses on Mistral
($0.45$ versus $0.40$). On held-out TEST, seven of the eight selected
model--compressor pairs have empirical material-degradation risk below
$5\%$; Mistral DefensiveKV has $61/1200$ events ($5.08\%$), one event
above that proportion. The TEST rates are descriptive and do not change
the CAL-based certification decision. In particular, the one-event excess
neither invalidates the CAL certificate nor by itself establishes that the
population risk exceeds $5\%$.

\begin{table}[t]
\centering
\footnotesize
\setlength{\tabcolsep}{4pt}
\begin{tabular}{lccc}
\toprule
Method & Retention & CAL risk & TEST risk \\
\midrule
\multicolumn{4}{l}{Llama-3.1-8B} \\
SnapKV            & .75 & 3.50\% & 2.50\% \\
AdaKV             & .65 & 3.33\% & 3.08\% \\
DefensiveKV       & .40 & 3.67\% & 4.08\% \\
Layer-DefensiveKV & .35 & 3.58\% & 4.00\% \\
\midrule
\multicolumn{4}{l}{Mistral-7B} \\
SnapKV            & .70 & 3.33\% & 3.33\% \\
AdaKV             & .70 & 3.58\% & 2.58\% \\
DefensiveKV       & .40 & 3.67\% & 5.08\% \\
Layer-DefensiveKV & .45 & 3.17\% & 3.92\% \\
\bottomrule
\end{tabular}
\caption{LongBench results under $(\tau,\epsilon,\delta)=(.10,.05,.05)$.
Retention is the CAL-selected retained-KV fraction; smaller values indicate
more aggressive compression. CAL and TEST columns show empirical
material-degradation event rates. Each model--compressor sequence is
calibrated separately.}
\label{tab:main_lb}
\end{table}

\begin{figure*}[t]
    \centering
    \includegraphics[width=0.72\textwidth]
    {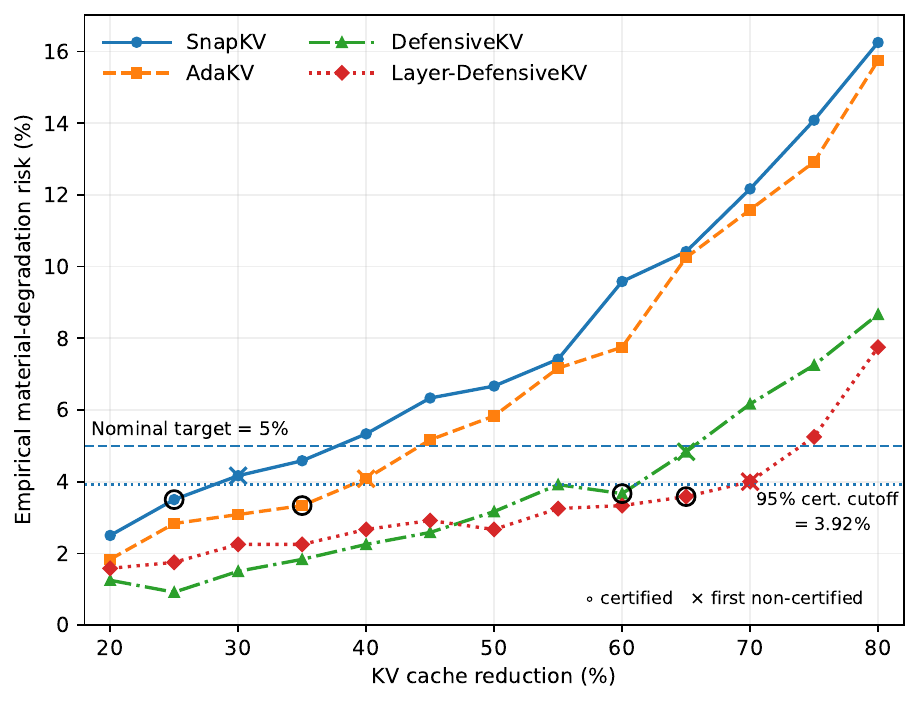}
    \caption{
    Empirical material-degradation risk for Llama-3.1-8B on LongBench CAL
    under $\tau=.10$. Horizontal lines denote the nominal risk target
    $\epsilon=.05$ and the finite-sample certification cutoff.
    Open circles and crosses indicate the certified and first
    non-certified operating points, respectively.
    }
    \label{fig:main_certification}
\end{figure*}

Figure~\ref{fig:main_certification} visualizes the full Llama-3.1-8B
calibration frontier. For every compressor, the next more
aggressive candidate fails certification despite having empirical risk below
the nominal 5\% target; Sec.~\ref{sec:finite_sample_results} analyzes these
first-fail cases in detail.

Across the tested grid, material-degradation risk generally increases as
retention decreases (Figure~\ref{fig:main_certification}). Retention is therefore a useful
empirical efficiency--reliability control variable, although the
certification procedure does not assume monotonicity. 

\subsubsection{Certified versus Empirical-Risk Selection}
\label{sec:finite_sample_results}

A naive plug-in rule would accept any candidate whose observed CAL risk
satisfies $\widehat{\risk}_{.10}\le .05$. With $n=1200$, this rule allows
up to 60 observed events. In contrast, under
$(\epsilon,\delta)=(.05,.05)$, the finite-sample test certifies at most
47 events.

The first non-certified candidate exceeds this cutoff for every eviction
method: SnapKV at 70\% retention has 50 violations (4.17\%), AdaKV at
60\% has 49 (4.08\%), DefensiveKV at 35\% has 58 (4.83\%), and
Layer-DefensiveKV at 30\% has 48 (4.00\%). Thus, all four candidates
would satisfy the nominal 5\% target under empirical thresholding, yet
none provides sufficient finite-sample evidence for certification under
the pre-specified level-$\delta=.05$ test. Across the four Llama
compressors, certification consequently selects retention ratios
$0.05$--$0.10$ higher than selection based on the observed CAL event rate
alone. For Layer-DefensiveKV, the plug-in rule selects retention $0.30$,
whereas certification selects $0.35$; on the same held-out TEST requests,
these policies yield $64/1200$ ($5.33\%$) and $48/1200$ ($4.00\%$)
material-degradation events, respectively. This comparison illustrates the difference between the selected operating points but does not establish a statistically significant paired difference or imply that the population risk at retention $0.30$ exceeds $5\%$ for future deployment requests. The TEST results do not affect certification.

\subsection{RULER-32K Results}
\label{sec:result_ruler}

To assess whether the certification behavior extends to controlled
long-context stress tests, we independently apply the same reliability
contract and fixed-sequence certification procedure to four eviction
methods on the task-balanced RULER-32K population. For each method, the
operating point is selected using only the RULER CAL split and then
evaluated on the common held-out TEST split; no LongBench-certified
retention ratio is transferred.

Table~\ref{tab:ruler_cert} summarizes the resulting operating points.
Layer-DefensiveKV permits the most aggressive compression, retaining
$0.35$ of the cache with CAL and TEST risks of $3.92\%$ and $3.54\%$,
respectively. DefensiveKV retains $0.45$ with a TEST risk of $3.00\%$,
whereas AdaKV requires $0.70$ retention and has a TEST risk of $2.77\%$.
SnapKV's first candidate, retention $0.80$, has $135/1300$ CAL events
($10.38\%$) and fails certification, so the procedure returns full KV. This fallback is specific to the evaluated sequence: retention fractions between $0.80$ and $1.00$ were not tested, so the result does not establish that every compressed SnapKV policy is infeasible.
Thus, all three certified compressed
policies remain empirically below the nominal $5\%$ target on held-out
TEST, while the selected retention varies substantially across eviction
methods. The aggregate result still masks task heterogeneity: TEST risk is
concentrated on common-word extraction (CWE), reaching $9\%$, $15\%$, and
$29\%$ for AdaKV,
DefensiveKV, and Layer-DefensiveKV, respectively. These per-task rates are
descriptive.

\begin{table}[t!]
\centering
\footnotesize
\setlength{\tabcolsep}{4pt}
\begin{tabular}{lccc}
\toprule
Method & Retention & CAL risk & TEST risk \\
\midrule
\multicolumn{4}{l}{Llama-3.1-8B} \\
SnapKV             & 1.00 & 0.00\% & 0.00\% \\
AdaKV              & .70 & 3.62\% & 2.77\% \\
DefensiveKV        & .45 & 3.77\% & 3.00\% \\
Layer-DefensiveKV  & .35 & 3.92\% & 3.54\% \\
\bottomrule
\end{tabular}
\caption{RULER-32K results under
$(\tau,\epsilon,\delta)=(.10,.05,.05)$.
Retention denotes the CAL-selected retained-cache fraction. SnapKV's first
compressed candidate fails certification, so its row reports the full-KV
fallback. CAL and TEST columns report empirical material-degradation event
rates.}
\label{tab:ruler_cert}
\end{table}

\subsection{Request-Adaptive Certification with ReFreeKV}
\label{sec:refree}

We examine request-adaptive KV-cache retention within the separate 8K
preprocessing protocol. ReFreeKV uses a global threshold but retains a
different KV fraction for each request. Under the pre-specified contract
$(\tau,\epsilon,\delta)=(.10,.05,.05)$, fixed-sequence LTT certifies
$T=.0005$ (39/1200 CAL events, 3.25\%); the next threshold, $T=.001$,
is not certified despite its 4.42\% empirical CAL risk. The selected
policy retains 87.48\% of the cache on average in CAL. On held-out TEST,
it retains 87.36\% and has 42/1200 (3.50\%) degradation events.

This result shows that the same risk-control framework can calibrate a global threshold for a request-adaptive compressor without prescribing a fixed retention ratio for every request.

\section{Additional Analyses}
\label{sec:additional}

The main experiments use the pre-specified contract
$(\tau,\epsilon,\delta)=(.10,.05,.05)$ across models and benchmarks. This
contract and the candidate order were fixed before the CAL outcomes used for
main certification were inspected. We subsequently recomputed alternative
choices for $\tau\in\{.05,.10\}$ and
$\epsilon\in\{.05,.10,.20\}$ from the saved Llama CAL utilities as an
exploratory sensitivity analysis. These post-hoc calculations do not alter
the frozen main operating points or their held-out TEST evaluations.

\begin{table}[t!]
\centering
\footnotesize
\setlength{\tabcolsep}{5pt}
\begin{tabular}{llccc}
\toprule
Compressor & $\tau$ & $\epsilon=.05$ & $\epsilon=.10$ & $\epsilon=.20$ \\
\midrule
\multirow{2}{*}{SnapKV}
& .05 & 1.00 & .60 & .30 \\
& .10 & .75 & .45 & .20 \\
\multirow{2}{*}{AdaKV}
& .05 & .80 & .55 & .30 \\
& .10 & .65 & .40 & .20 \\
\multirow{2}{*}{DefensiveKV}
& .05 & .65 & .35 & .20 \\
& .10 & .40 & .25 & .20 \\
\multirow{2}{*}{Layer-DefensiveKV}
& .05 & .65 & .30 & .20 \\
& .10 & .35 & .20 & .20 \\
\bottomrule
\end{tabular}
\caption{Exploratory Llama LongBench contract sensitivity recomputed from
the same saved request-level CAL utilities. The retained-KV
fraction selected by applying fixed-sequence testing separately at
$\delta=.05$ for each contract is reported; they do not constitute a simultaneous
guarantee across contracts. The pre-specified main contract is
$(\tau,\epsilon)=(.10,.05)$. A value of $1.00$ denotes full-KV fallback,
and $.20$ is the lowest retained fraction in the tested grid.}
\label{tab:contract_frontier}
\end{table}

\paragraph{Sensitivity to the reliability contract.}
Table~\ref{tab:contract_frontier} shows the CAL-selected retention for
$\tau\in\{.05,.10\}$ and $\epsilon\in\{.05,.10,.20\}$, using $\delta=.05$
separately for each contract. In prospective use, a deployment owner specifies the tolerated loss $\tau$
and its allowed frequency $\epsilon$ before inspecting calibration outcomes,
potentially using a separate development (DEV) set; the procedure then
returns a retained fraction or the full-KV fallback.

At $\epsilon=.05$, tightening $\tau$ from $.10$ to $.05$ raises
Layer-DefensiveKV retention from $.35$ to $.65$ and moves SnapKV from
$.75$ to full KV. At $\epsilon=.20$, six of eight entries reach $.20$,
the lowest ratio evaluated. For fixed $\tau$, increasing $\epsilon$ raises
the nominal risk target in Fig.~\ref{fig:main_certification} and also changes
the finite-sample certification cutoff.

Except for the pre-specified main contract, these post-hoc cells are
descriptive. A nominal $\delta=.05$ interpretation would apply prospectively
to a contract fixed independently of CAL. Choosing a favorable contract after
inspecting the table requires fresh calibration data or an appropriate
multiplicity correction.

\begin{table*}[t]
\centering
\scriptsize
\setlength{\tabcolsep}{1.5pt}
\resizebox{\textwidth}{!}{%
\begin{tabular}{ll*{13}{c}}
\toprule
Llama-3.1-8B & & \multicolumn{3}{c}{Multi-Doc QA} & \multicolumn{2}{c}{Summarization} & \multicolumn{3}{c}{Few-shot Learning} & \multicolumn{2}{c}{Synthetic} & \multicolumn{2}{c}{Code} & \\
\cmidrule(lr){3-5}\cmidrule(lr){6-7}\cmidrule(lr){8-10}\cmidrule(lr){11-12}\cmidrule(lr){13-14}
Method & & \rotatebox[origin=l]{30}{HotpotQA} & \rotatebox[origin=l]{30}{2WikiMQA} & \rotatebox[origin=l]{30}{MuSiQue} & \rotatebox[origin=l]{30}{GovReport} & \rotatebox[origin=l]{30}{MultiNews} & \rotatebox[origin=l]{30}{TREC} & \rotatebox[origin=l]{30}{TriviaQA} & \rotatebox[origin=l]{30}{SAMSum} & \rotatebox[origin=l]{30}{PassageCount} & \rotatebox[origin=l]{30}{PassageRetrieval-en} & \rotatebox[origin=l]{30}{LCC} & \rotatebox[origin=l]{30}{RepoBench-P} & Avg. \\
\midrule
\multirow{2}{*}{Full KV} & mean & 58.7 & 49.0 & 29.8 & 35.0 & 27.8 & 72.0 & 96.4 & 44.0 & 8.0 & 99.0 & 71.0 & 52.2 & 53.6 \\
 & risk & -- & -- & -- & -- & -- & -- & -- & -- & -- & -- & -- & -- & -- \\
\addlinespace[2pt]
\multirow{2}{*}{SnapKV (.75)} & mean & 58.8 & 49.5 & 29.1 & 34.0 & 27.1 & 69.0 & 96.1 & 44.2 & 6.0 & 99.0 & 72.3 & 52.7 & 53.2 \\
 & risk & 1\% & 6\% & 5\% & 1\% & 1\% & 4\% & 1\% & 2\% & 2\% & 0\% & 1\% & 6\% & 2.50\% \\
\addlinespace[2pt]
\multirow{2}{*}{AdaKV (.65)} & mean & 58.2 & 54.0 & 27.9 & 33.1 & 26.6 & 69.0 & 95.1 & 43.7 & 5.3 & 99.0 & 71.8 & 54.2 & 53.2 \\
 & risk & 3\% & 3\% & 6\% & 2\% & 1\% & 4\% & 2\% & 4\% & 3\% & 0\% & 2\% & 7\% & 3.08\% \\
\addlinespace[2pt]
\multirow{2}{*}{DefensiveKV (.40)} & mean & 57.0 & 48.7 & 27.8 & 33.5 & 26.8 & 70.0 & 96.4 & 43.6 & 7.0 & 99.0 & 72.3 & 56.3 & 53.2 \\
 & risk & 5\% & 10\% & 9\% & 0\% & 3\% & 2\% & 0\% & 8\% & 2\% & 0\% & 4\% & 6\% & 4.08\% \\
\addlinespace[2pt]
\multirow{2}{*}{Layer-DefensiveKV (.35)} & mean & 57.2 & 48.8 & 26.2 & 33.7 & 27.0 & 70.0 & 95.9 & 43.9 & 6.0 & 100.0 & 75.2 & 58.4 & 53.5 \\
 & risk & 3\% & 8\% & 8\% & 2\% & 2\% & 2\% & 2\% & 10\% & 3\% & 0\% & 3\% & 5\% & 4.00\% \\
\midrule
\multicolumn{15}{l}{\emph{ReFreeKV protocol (8K preprocessing and separate prompt formatting)}} \\
\multirow{2}{*}{Full KV}
 & mean & 18.0 & 42.7 & 6.9 & 34.8 & 27.7 & 12.0 & 82.1 & 17.1 & 4.1 & 74.0 & 26.1 & 39.2 & 32.1 \\
 & risk & -- & -- & -- & -- & -- & -- & -- & -- & -- & -- & -- & -- & -- \\
\addlinespace[2pt]
\multirow{3}{*}{ReFreeKV ($T=.0005$)}
 & retention & 87.5\% & 85.3\% & 87.5\% & 94.8\% & 97.7\% & 94.3\% & 75.7\% & 93.4\% & 82.9\% & 87.0\% & 85.6\% & 76.7\% & 87.36\% \\
 & mean & 21.0 & 43.7 & 7.4 & 34.2 & 27.6 & 19.1 & 71.9 & 17.4 & 3.4 & 74.2 & 26.3 & 37.5 & 32.0 \\
 & risk & 1\% & 0\% & 2\% & 1\% & 0\% & 0\% & 20\% & 1\% & 2\% & 4\% & 3\% & 8\% & 3.50\% \\
\bottomrule
\end{tabular}%
}
\caption{LongBench task-level TEST results at CAL-selected
settings. Mean utility is on the 0--100 scale; risk is the
material-degradation event rate (\%) at $\tau=.10$. Fixed-budget retention
is shown in parentheses, while the ReFreeKV retention row reports mean
retained KV (\%). Each task has 100 TEST requests; per-task risks are
descriptive.}
\label{tab:task_heterogeneity_llama}
\end{table*}

\paragraph{Task and metric heterogeneity.}
Table~\ref{tab:task_heterogeneity_llama} reports held-out TEST outcomes
for twelve LongBench tasks at the CAL-selected settings of four fixed-budget
compressors and ReFreeKV. The fixed-budget methods have aggregate risks of
$2.50\%$--$4.08\%$, yet DefensiveKV reaches $10\%$ on 2WikiMQA and
Layer-DefensiveKV reaches $10\%$ on SAMSum. ReFreeKV has $3.50\%$ aggregate
risk but $20\%$ on TriviaQA; its mean retention ranges from $75.7\%$ on
TriviaQA to $97.7\%$ on MultiNews. Under its separate evaluation protocol,
ReFreeKV's aggregate mean utility changes only from $32.1$ under full KV to
$32.0$. Thus, adaptive retention and stable aggregate utility need not imply
uniform task-level reliability.

These task-level rates are empirical estimates from 100 TEST requests per
task; the certificate applies to the task-balanced mixture and does not
provide a per-task $5\%$ guarantee. To assess whether the largest observed
task-level risks reflect sampling variability or task-specific population
risks exceeding $\epsilon$, we conduct a post-hoc simultaneous analysis of
held-out TEST outcomes for all 147 nontrivial selected-policy--task
combinations: 60 from Llama LongBench, 48 from Mistral LongBench, and 39
from RULER. Combinations with full-KV fallback are excluded because their
degradation risk is zero by definition. We construct two-sided
Clopper--Pearson intervals with a Bonferroni correction across the full
family. Even under this correction, the lower confidence bound exceeds
$.05$ for RULER common-word extraction under Layer-DefensiveKV
($29/100$; 95\% simultaneous CI: $[14.6\%,47.1\%]$) and DefensiveKV
($15/100$; $[5.1\%,31.1\%]$), and for LongBench TriviaQA under ReFreeKV
($20/100$; $[8.2\%,37.1\%]$). These results provide evidence that some
task-specific population risks exceed $5\%$ even when the policy is
certified for the task-balanced mixture. The ReFreeKV result further
suggests that request adaptivity alone may be insufficient for uniform
task-level reliability, motivating risk-aware adaptive policies that
allocate KV retention according to estimated material-degradation risk.

Task means and degradation events capture different properties. On LCC,
full-KV utility in the fixed-budget protocol averages $71.0$ points and
Layer-DefensiveKV at retention $0.35$ averages $75.2$ points, yet $3\%$ of
requests lose more than ten points relative to full KV. Because LongBench
tasks use different example-level metrics, the common threshold $\tau=.10$
consistently represents a ten-point utility loss on the 0--100 scale but has
task-dependent semantic meaning and need not correspond to a
correct-to-incorrect transition.

In the fixed-budget protocol, full-KV utility is zero on $253/1200$ TEST
requests. More generally, because compressed utility is nonnegative, any
request with $U_{\mathrm{full}}(X)\le\tau$ cannot incur degradation strictly
greater than $\tau$. Low relative-degradation risk therefore does not imply
high absolute task quality and must be interpreted alongside the full-KV
reference utility.

\section{Limitations}
\label{sec:limitations}

The certificate is specific to the declared benchmark population and assumes independent calibration requests drawn from each task. It does not imply task-specific guarantees or robustness under distribution shift. Correlated samples or simultaneous guarantees across tasks would require an appropriate sampling and calibration design.

We evaluate two 7--8B model families and a limited set of eviction methods
on discrete candidate grids. The selected operating points may not transfer
to larger models, other architectures, or serving stacks. Finer grids could
improve resolution but would increase the offline cost of paired full-KV and
candidate evaluation.

The guarantee requires the contract, candidate policies, and their order to
be fixed before CAL outcomes are inspected; post-hoc selection requires
multiplicity control or fresh calibration data. The choices of $\tau$ and
$\epsilon$ are application-dependent, and because degradation is defined
relative to full KV, low certified risk does not guarantee high absolute
utility. We do not directly evaluate system-level latency or throughput; our analysis characterizes retained-KV fraction and certifies utility risk rather than end-to-end serving efficiency.

\section{Conclusion}
\label{sec:conclusion}

We formulate KV-cache eviction as a reliability-first retention-selection
problem: rather than fixing a memory budget and evaluating average quality
afterward, we specify an allowable frequency of material request-level
degradation and use calibration data to select a retention policy. Building
on fixed-sequence Learn-then-Test (LTT), the resulting procedure provides
finite-sample risk control for the returned policy without modifying the underlying compressor or assuming
monotone risk. 

Across multiple eviction methods, models, and long-context benchmarks, the
same reliability contract leads to substantially different retention
choices. We further show that average utility can conceal request-level
failures and that policies meeting the empirical calibration-risk target can
still fail finite-sample certification. The framework therefore shifts
KV-cache compression from budget-first evaluation toward risk-controlled
deployment, providing a principled way to determine how aggressively KV
memory can be reduced for a declared population and utility metric.

\bibliography{risk_kv_tacl}
\bibliographystyle{acl_natbib}

\end{document}